%% file: MOE_Paper.tex
\documentclass[10pt,twocolumn,letterpaper]{article}

\usepackage{iccv}
\usepackage{times}
\usepackage{epsfig}
\usepackage{graphicx}
\usepackage{amsmath}
\usepackage{amssymb}

\usepackage[breaklinks=true,bookmarks=false,colorlinks,citecolor=green,linkcolor=green]{hyperref}

\iccvfinalcopy % *** Uncomment this line for the final submission

\def\iccvPaperID{3894} % *** Enter the ICCV Paper ID here

\ificcvfinal\fi

\usepackage[nolist]{acronym} % Acronyms from file
\input{acronyms} % Self-created acronyms
\usepackage{booktabs} % Nice looking tables
\usepackage{multirow}

\usepackage[dvipsnames]{xcolor}

\newcommand{\hmm}[1]{{#1}}

\usepackage{caption}
\usepackage{subcaption}
\usepackage{dblfloatfix}

\begin{document}

% \title{Mixture of Motion Primitives: \\Sign Language Production via Translation and Animation}

\title{Mixed SIGNals: Sign Language Production via a Mixture of Motion Primitives}

\author{Ben Saunders, Necati Cihan Camgoz, Richard Bowden\\
University of Surrey\\
{\tt\small \{b.saunders, n.camgoz, r.bowden\}@surrey.ac.uk}
}
\maketitle
% Remove page # from the first page of camera-ready.
\ificcvfinal\thispagestyle{empty}\fi

\begin{abstract}
    It is common practice to represent spoken languages at their phonetic level. However, for sign languages, this implies breaking motion into its constituent motion primitives. Avatar based \acf{slp} has traditionally done just this, building up animation from sequences of hand motions, shapes and facial expressions. However, more recent deep learning based solutions to \ac{slp} have tackled the problem using a single network that estimates the full skeletal structure.
   
    We propose splitting the \ac{slp} task into two distinct jointly-trained sub-tasks. The first translation sub-task translates from spoken language to a latent sign language representation, with gloss supervision. Subsequently, the animation sub-task aims to produce expressive sign language sequences that closely resemble the learnt spatio-temporal representation. Using a progressive transformer for the translation sub-task, we propose a novel \acf{momp} architecture for sign language animation. A set of distinct motion primitives are learnt during training, that can be temporally combined at inference to animate continuous sign language sequences.
    
    We evaluate on the challenging \ac{ph14t} dataset, presenting extensive ablation studies and showing that \ac{momp} outperforms baselines in user evaluations. We achieve state-of-the-art back translation performance with an 11\% improvement over competing results. Importantly, and for the first time, we showcase stronger performance for a full translation pipeline going from spoken language to sign, than from gloss to sign.
\end{abstract}

\begin{figure}[t!]
    \centering
    \includegraphics[width=1.00\linewidth]{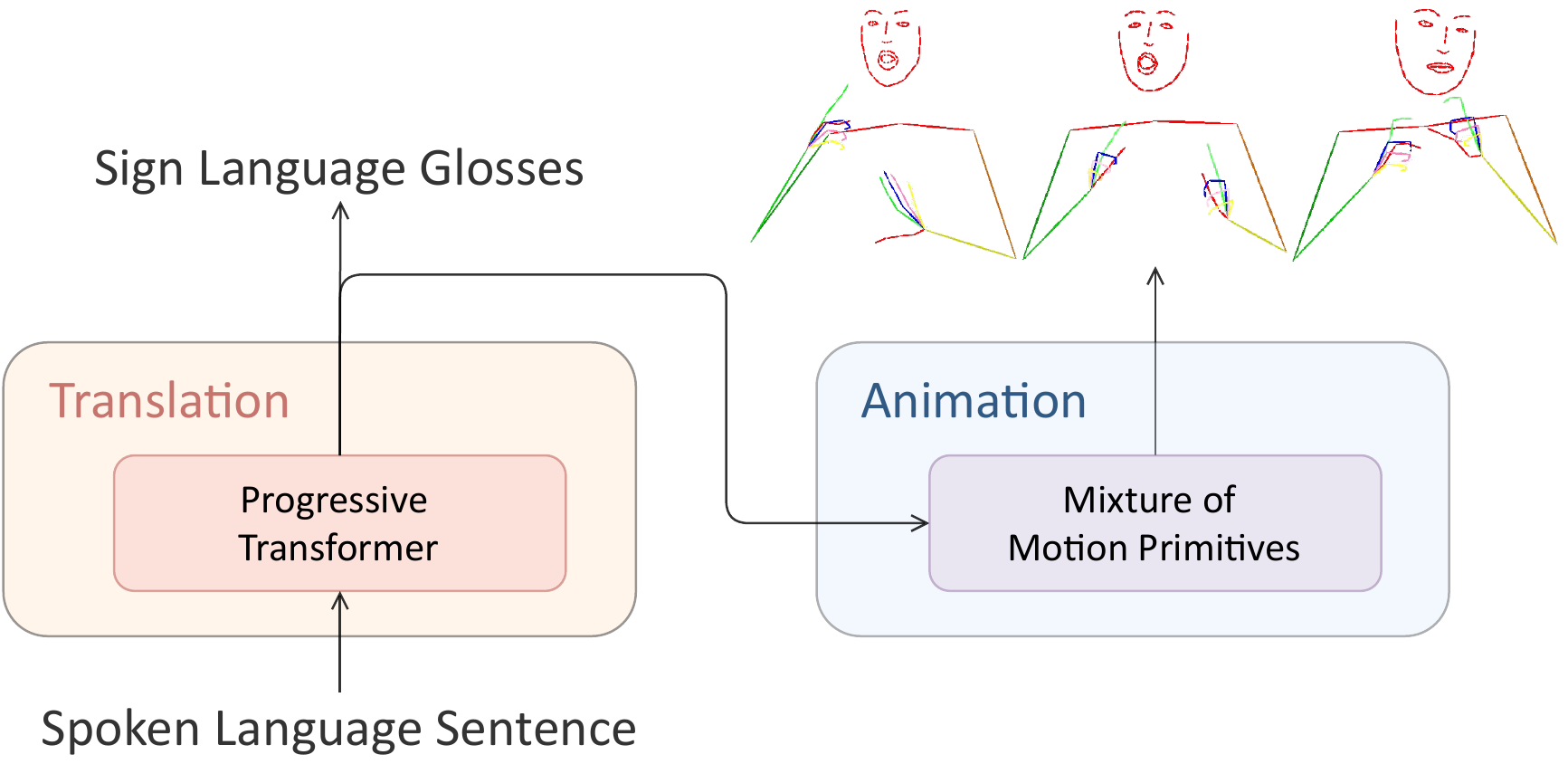}
    % \draftfig
    \caption{An overview of the proposed \ac{slp} sub-tasks of Translation and Animation.}
    \label{fig:translation_animation}
\end{figure}%

%%%%%%%%% BODY TEXT
\section{Introduction} \label{sec:introduction}

% Sign Language can be made up of Motion Primitives
Sign languages are visual languages used for communication by the Deaf communities. Analogous to phonemes in speech, sign languages can be broken down into cheremes, the smallest distinctive structural units \cite{stokoe1980sign}. Cheremes can be represented as motion primitives, a set of manual and non-manual motions\footnote{Manual features are the hand shape and motion, whereas non-manuals are the mouthings and facial expressions} that are combined to represent all sign language utterances. Such phonetic representations are typically used by linguists for annotation \cite{hankehamnosys,sutton2014lessons} or in graphical based avatars for sign generation.

% Why is SLP in a single task bad
\acf{slp}, the translation from spoken language sentences to continuous sign language sequences, requires both an accurate translation and expressive animation. Previous work has traditionally tackled these two sub-task as a single task with one network \cite{saunders2020progressive,stoll2018sign,zelinka2020neural}, leading to under-expressive production. Although previous \ac{slp} models have used gloss\footnote{Glosses are a written representation of sign, defined as minimal lexical items.} as an intermediate representation \cite{stoll2020text2sign}, this creates an information bottleneck that disregards the contextual information available in the original text.

% Introduce MoMP and the SLP sub-tasks of Translation and Animation 
In this paper, we propose to formulate \ac{slp} as two distinct but jointly-trained sub-tasks as can be seen in Figure \ref{fig:translation_animation}: 1) An initial translation from spoken language to a sign language representation, with gloss supervision; 2) Subsequent animation of a visual sign language sequence. This can be viewed as analogous to a text-to-speech pipeline with an initial translation into phonemes and a subsequent vocalisation. However, we do not force a gloss information bottleneck but instead condition learning on gloss, resulting in significant performance increases. 

% Mixture of Experts model
We utilise a progressive transformer model as our translation backbone \cite{saunders2020progressive}. Sign language representations are learnt per frame using the gloss supervision. This prompts the sub-network to learn meaningful representations for the end goal of sign language production.

To animate expressive sign language sequences, we propose a novel \acf{momp} network. Based on a \ac{moe} architecture, we learn a combination of distinct motion primitives that are able to produce an infinite number of unique sign language utterances. Due to the continuous nature of sign language, we apply expert blending on a per frame basis, thus enabling different experts to be activated for separate sections of the output sequence.

As the subset of motion primitives is significantly smaller than the full set of signs, the animation sub-task is reduced to a gating network that selects the correct primitive to animate for specific sections of the full sequence. This also enables a scaling of \ac{slp} models to larger datasets, with new signs being a novel combination of the learnt primitives. We represent experts as masked transformer encoders, using self-attention to learn unique structural motions. We use a further transformer encoder model for the gating network, thus building, to the best of our knowledge, the first full transformer-based \ac{moe} architecture.

% Experiments
We evaluate on the challenging \ac{ph14t} corpus, performing an extensive ablation study of the proposed network and conducting a user evaluation that highlights the animation quality of the \ac{momp}. Furthermore, we achieve state-of-the-art \ac{slp} back translation results and showcase, for the first time, stronger performance for a full translation pipeline that produces sign sequences directly from the source spoken language, than from an intermediate gloss representation.

The contributions of this paper can be summarised as:
\begin{itemize} \itemsep0em
    \item A novel transformer-based \ac{moe} architecture, \acf{momp}, that combines learnt motion primitives at the frame level.
    \item The first \ac{slp} model to separately model the sub-tasks of translation and animation.
    \item State-of-the-art \ac{slp} performance and user evaluation results on the \ac{ph14t} dataset.
    \item The first \ac{slp} model to achieve higher performance for a full translation pipeline going from spoken language to sign, than from gloss to sign.
\end{itemize}

% Structure of Paper
The rest of this paper is organised as follows: In Section~\ref{sec:related_work}, we review the literature in \ac{slp} and \ac{moe}. In Section~\ref{sec:methodology}, we outline the proposed \ac{momp} network. We present quantitative and qualitative model comparison in Section~\ref{sec:experiments}, and finally conclude in Section \ref{sec:conclusion}.

\section{Related Work} \label{sec:related_work}

\paragraph{Sign Language Motion Primitives}

Phonemes are defined as the smallest distinctive structural units of spoken language that can be combined to create an infinite number of meaningful utterances \cite{lass1984phonology,trubetzkoy1969principles}. Cheremes are used as the equivalent representation specific to sign language \cite{brentari1998prosodic,corina1993nature,stokoe1980sign}. This phonetic structure of sign language includes the sublexical parameters of shape, movement and location used to describe the motion and structure of all signs \cite{brentari2002modality,fenlon2018phonology}. Motion primitives can be seen as a subset of cheremes, encompassing the gestural motions of both manual and non-manual features. Although the possible motion primitives are much smaller in number than the full set of signs, they can be combined to recreate all unique sign language sequences.

\paragraph{Sign Language Production}

Computational sign language research has been prominent for the last 30 years \cite{bauer2000video,starner1997real,tamura1988recognition}. Previous research has focused on isolated sign recognition \cite{antonakos2012unsupervised,grobel1997isolated,ozdemir2016isolated}, \ac{cslr} \cite{camgoz2017subunets,cui2017recurrent,koller2019weakly,koller2020quantitative} and, more recently, the task of \ac{slt} \cite{camgoz2018neural,camgoz2020multi,ko2019neural,orbay2020neural}. Camgoz \etal \cite{camgoz2020sign} proposed a jointly trained \ac{cslr} and \ac{slt} system, showing a performance increase for both tasks.

\acf{slp}, the translation from spoken language sentence to sign language sequences, has traditionally been tackled using avatars \cite{bangham2000virtual,cox2002tessa,ebling2015bridging,elliott2008linguistic,glauert2006vanessa}. Animating sign using avatars helps to separate the translation task from the animation, with an initial manual translation from text into a sign language representation such as HamNoSys \cite{hankehamnosys} or SignWriting \cite{sutton2014lessons}.

In contrast, more recent works have applied deep learning to \ac{slp} \cite{cui2019deep,miyazaki2020machine,saunders2020adversarial,saunders2020everybody,saunders2021anonysign,stoll2018sign,xiao2020skeleton,zelinka2020neural}, with Saunders \etal \cite{saunders2020progressive} proposing the first \ac{slp} model to translate from spoken language sentences to sign language sequences in an end-to-end manner. However, these methods combine both the translation and animation elements into a single pipeline, leading to a lack of expressive animation. Stoll \etal \cite{stoll2020text2sign} use the intermediate representation of gloss but this creates an information bottleneck that all sign must pass through. In this work, we separate the animation from the translation sub-task using a joint supervision of both gloss and skeletal pose. Furthermore, we combine a learnt set of motion primitives that can animate any sign language utterance and use gloss to condition learning rather than form a bottleneck.

\begin{figure*}[t!]
    \centering
    \includegraphics[width=0.95\linewidth]{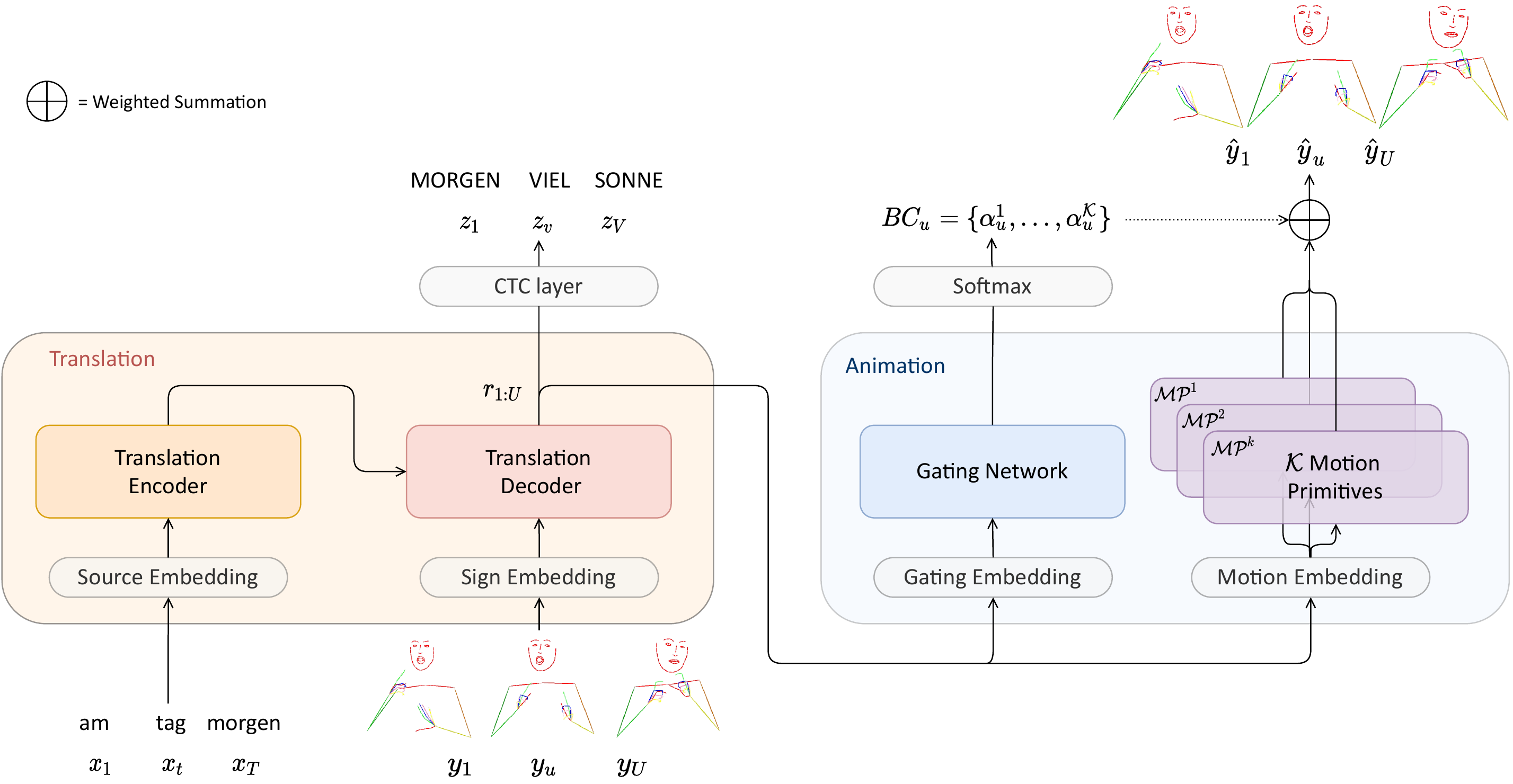}
    \caption{\acf{momp} network overview, showing an initial translation sub-task from spoken language, $x_{1:T}$, to sign language representation with gloss supervision, $z_{1:V}$ (left). A subsequent animation sub-task uses a blended mixture of $\mathcal{K}$ learnt motion primitives, $\mathcal{MP}^{\mathcal{K}}$, to produce a continuous sign language sequence, $\hat{y}_{1:U}$, (right).}
    \label{fig:MOMP_model_overview}
\end{figure*}%
\paragraph{Mixture of Experts}

\acfp{moe} are a jointly-trained ensemble of expert systems, each locally specialised for a different subdomain of inputs \cite{jacobs1991adaptive,jordan1994hierarchical}. A gating network predicts a set of blending coefficients, used to weight the decision of each expert in the final output. We refer to Gormley \etal for a broad survey on \acp{moe} \cite{gormley2019mixtures}.

Recently, \acp{moe} have become popular in \ac{nmt} \cite{garmash2016ensemble,peng2020mixture,ramachandran2018diversity,yang2017breaking}. Sparsely gated \acp{moe} use a \ac{moe} layer to enable extensive scaling of parameters, requiring only a subset of experts to be computed for each sequence \cite{fedus2021switch,lepikhin2020gshard,ryabinin2020towards,shazeer2017outrageously}. \acp{moe} have also been used to promote diversity in text generation \cite{cho2019mixture,he2018sequence,shen2019mixture} and even to enable multi-task learning \cite{kaiser2017one,ma2018modeling}.

Stoll \etal \cite{stoll2020signsynth} applied a \acp{moe} concept to \ac{slp}, producing isolated signs from a small vocabulary of 105 glosses. Our work produces continuous sign sequences directly from spoken language, using distinct motion primitives combined at the output level for 1066 glosses.

Combining transformers with \acp{moe} is motivated by research that suggests transformer networks are overparameterised \cite{michel2019sixteen,voita2019analyzing}. Peng \etal \cite{peng2020mixture} build experts consisting of multiple transformer heads and Lepikhin \etal \cite{lepikhin2020gshard} build large-scale \ac{nmt} models using transformer \acp{moe}. Our work differs in that we represent each expert as an individual transformer encoder and propose a transformer-based gating network. Additionally, we apply \acp{moe} at the token i.e. frame level as opposed to the sequence level, enabling the modelling of specialised motion primitives.

Probably closest to our work is the approach of Zhang \etal \cite{zhang2018mode}, who use \acp{moe} to model repetitive gait sequences of quadraped motion. Each expert is trained to be specialised in the production of a certain type of motion. However, we tackle the subtle motions of sign language in the context of translation and, to achieve the required subtleties, perform blending at the output level rather than in the feature space.

\section{Methodology} \label{sec:methodology}

Given a source spoken language sequence, \hbox{$\mathcal{X} = (x_{1},...,x_{T})$} with $T$ words, the objective of an \ac{slp} model is to produce a sign language sequence, \hbox{$\mathcal{Y} = (y_{1},...,y_{U})$} with $U$ frames. State-of-the-art \ac{slp} works have approached the task using a single end-to-end network with no intermediate representation \cite{saunders2020progressive,zelinka2020neural}. This simultaneously tackles the challenging tasks of accurate translation into sign language grammar and expressive animation of sign language motion with a single unified loss function, impacting the networks ability to perform well at either task.

Motivated by this, we propose to split the \ac{slp} task into two jointly-trained sub-tasks: 1) An initial translation task from spoken language to sign language representation, with gloss supervision, \hbox{$\mathcal{Z} = (z_{1},...,z_{V})$} with $V$ glosses; 2) A subsequent sign language animation task in the form of skeletal pose sequences. We propose a \acf{momp} network that employs a progressive transformer for sign language translation (Left of Figure \ref{fig:MOMP_model_overview}) and a novel \ac{moe} architecture for sign language production (Right of Figure \ref{fig:MOMP_model_overview}). In the remainder of this section we describe each component of \ac{momp} in detail.
%, with an overview provided in Figure \ref{fig:MOMP_model_overview}.

\subsection{Translation: Progressive Transformer}

As shown in Figure \ref{fig:MOMP_model_overview}, we utilise a progressive transformer network \cite{saunders2020progressive} for the translation sub-task, which learns to translate from spoken language to sign language representation. A transformer encoder learns a representation of the input spoken language sentence, $x_{1:T}$, to pass to the auto-regressive transformer decoder. Given a sign language sequence, $y_{1:U}$, and the respective counter values, the decoder learns a sign language representation on a per frame basis, $\mathcal{R} = r_{1:U}$:
\begin{equation}
    r_{u+1} = \mathrm{Translation}(y_{u} |y_{i:u-1}, \mathcal{X})
\end{equation}

Then, using a separate linear layer, we provide the translation sub-network with additional supervision from gloss information during training, prompting the model to learn a meaningful latent temporal representation for the ultimate goal of sign language production. Due to the lack of frame-level gloss annotation, we use a \ac{ctc} layer to provide supervision in a sequence-to-sequence manner \cite{graves2006connectionist}. The \ac{ctc} layer uses the decoded latent representations for each frame, $r_{1:U}$, and computes $p(\mathcal{Z}|\mathcal{R})$ by marginalising over all possible alignments:
\begin{equation}
    p(\mathcal{Z}|\mathcal{R})   = \sum_{\pi \in \mathcal{B}} p(\mathcal{\pi}|\mathcal{R})
\end{equation}
where $\pi$ is a path and $\mathcal{B}$ are the set of all viable paths that correspond to $\mathcal{Z}$. The translation loss can then be calculated as:
\begin{equation}
    \mathcal{L}_{\mathcal{T}} = 1 - p(\mathcal{Z}^{*}|\mathcal{R})
\end{equation}
where $\mathcal{Z}^{*}$ is the ground truth gloss sequence.

\subsection{Animation: Mixture of Motion Primitives}

To produce expressive sign poses from the translated sign language representation, we learn a \acf{momp} network (right of Figure \ref{fig:MOMP_model_overview}) that combines learnt motion primitives at the frame-level using \acp{moe}. \acp{moe} are a common technique for ensemble learning \cite{jacobs1991adaptive} where an ensemble of $\mathcal{K}$ expert systems, $\{\mathcal{MP}^{i}\}^{k}_{i=1}$, are jointly trained, each locally specialised across different domains of expertise to produce an output $y_{u+1}^{k} = \mathcal{MP}^{k}(r_{u})$. 

A gating network, $GN$, learns a set of blending coefficients, $BC_{u} = \{\alpha_{u}^{1},...,\alpha_{u}^{k}\}$, used to weight the decision of each expert in the final output. Contrary to traditional \ac{moe} architectures that apply a unique blend per sequence \cite{peng2020mixture,shazeer2017outrageously,shen2019mixture}, we generate unique blending coefficients for each frame of the output sequence. This enables distinct motion primitives to be learnt for certain sections of the output sign language sequence.

\paragraph{Gating Network}

We utilise a transformer encoder with subsequence masking for the gating network, $GN$, using self-attention to learn the correct expert allocation. We mask future time-steps as in the transformer decoder, to disable the view of the future. Formally, the gating network produces a set of blending coefficients, $BC_{u+1}$, conditioned on the translated sign language representation, $r_{u}$:
\begin{equation}
    BC_{u+1} = \{\alpha_{u+1}^{1},...,\alpha_{u+1}^{\mathcal{K}}\} = GN(r_{u}|r_{1:u-1})
\end{equation}
where a softmax operation is applied to the produced blending coefficients to ensure $\alpha_{u}^{k} > 0$ and $\sum_{i=0}^{k} \alpha_{u}^{k} = 1$.

\paragraph{Motion Primitives}

A continuous sign language sequence consists of multiple distinct sections of motion. For example, a hand moving up then subsequently across the body. Our goal is to represent each of these distinct multi-frame motions as separate motion primitives, which can be temporally combined to produce a full sequence of uninterrupted, continuous signing motion. During training, each learnt motion primitive is encouraged to account for separate sections of the data, becoming specialised for specific motions that can be stitched together at inference.

Similar to the gating network, we build motion primitives, $\mathcal{MP}^{k}$, using transformer encoders with subsequence masking. We use self-attention over the translated sign language representation to learn the desired motion. We avoid conditioning on the source spoken language to ensure this sub-task is solely focused on animation. Formally, the output of each motion primitive can be computed as:
\begin{equation}
    y^{k}_{u+1} = \mathcal{MP}^{k}(r_{u})
\end{equation}

\noindent Each output frame is therefore a sign pose, $y_{u+1}$ produced in an auto-regressive manner by blending motion primitives, $y^{k}_{u+1}$, with their respective blending coefficients, $\alpha_{u+1}^{k}$:
\begin{equation}
    y_{u+1} = \mathrm{Animation}(r_{u}|r_{1:u-1}) = \sum_{i=1}^{\mathcal{K}} \alpha_{u+1}^{k} y^{k}_{u+1}
\end{equation}
with $\mathcal{K}$ experts and $\sum_{i=1}^{\mathcal{K}} \alpha_{u+1}^{k} = 1$. \hmm{As in \cite{saunders2020progressive}, the respective counter value is also produced for each frame.} Once the full sign pose sequence is produced, the animation loss, $\mathcal{L}_{\mathcal{A}}$ is calculated as the \ac{mse} loss between the predicted, $\hat{y}_{1:U}$, and ground truth, $y_{1:U}^{*}$ sequences:
\begin{equation}
    \mathcal{L}_{\mathcal{A}} = \frac{1}{U} \sum_{i=1}^{u} ( y_{1:U}^{*} - \hat{y}_{1:U} ) ^{2}
\end{equation}

We train our network by minimising the overall \ac{slp} loss, $\mathcal{L}_{\mathcal{SLP}}$, which is a weighted sum of the CTC based translation loss, $\mathcal{L}_{T}$, and the joint distance animation loss, $\mathcal{L}_{A}$, as:
\begin{equation}
    \mathcal{L}_{\mathcal{SLP}} = \lambda_{T} \mathcal{L}_{\mathcal{T}} + \lambda_{A} \mathcal{L}_{\mathcal{A}}
\end{equation}
where $\lambda_{T}$ and $\lambda_{A}$ weight the importance of each loss function during training and are evaluated in section \ref{sec:translation_animation_evaluation}.

\subsection{Training Schedule}

Naive end-to-end training of an \ac{moe} with backpropagation has been shown to result in a degenerate local minimum where expert weightings are consistent regardless of the input \cite{peng2020mixture,shen2019mixture}. Therefore, we use a combination of \acf{bcd} training and expert balancing losses to overcome this phenomenon, as described below.

\paragraph{Block Coordinate Descent} \label{sec:bcd}

We apply a \ac{bcd} training schedule, as introduced by Peng \etal \cite{peng2020mixture}, that decomposes updates into two interleaving steps $\mathcal{G}$ and $\mathcal{F}$: The $\mathcal{G}$ step processes a forward pass with blended outputs, fixes the translation sub-network and motion primitives and updates \textit{only} the gating network, $GN$; The $\mathcal{F}$ step then freezes the gating network and updates the full translation sub-network alongside a single expert for each frame, $E_{k}(x)$, sampled from the blending coefficient weights. During training, $\mathcal{G}$ steps are required less often than $\mathcal{F}$ steps, with a ratio of 3 $\mathcal{F}$ steps for each $\mathcal{G}$ step achieving best performance. 

Specific to \acp{moe}, \ac{bcd} forces a specialisation of experts for particular sections of sign pose sequences, learning the important motion primitives. Motivated by the comparison to dropout \cite{peng2020mixture}, we add a random chance of selecting expert $k$, with an annealing of the probability during training.

\paragraph{Expert Balancing Losses}

As seen in previous \ac{moe} architectures \cite{eigen2013learning,shazeer2017outrageously,shen2019mixture}, we find that a small subset of experts tend to be imbalanced and receive higher blending coefficients. This effect is self-reinforcing as the popular experts are trained quicker, receiving further allocation. In addition, as \ac{momp} produces continuous sign pose sequences with blending coefficients applied per frame, we favour gating networks with sparse activations. This avoids a weighted average of two motion primitives that may itself not be valid.

Following Bengio \etal \cite{bengio2015conditional}, we take a soft constraint approach to expert balancing and apply two regularisation terms. The first term is a balancing loss, $\mathcal{L}_{\mathcal{B}}$, that encourages an equal expert share in expectation:
\begin{equation}
    \mathcal{L}_{\mathcal{B}} =  \sum_{u=1}^{U} \frac{1}{\mathcal{K}} \sum_{k=1}^{\mathcal{K}}  (\alpha_{u}^{k} - \tau)^2
\end{equation}
where $\tau$ is the expected balanced load, $\frac{1}{\mathcal{K}}$. 

The second term is a variance loss, $\mathcal{L}_{\mathcal{V}}$, that encourages a sparse allocation per frame:
\begin{equation}
    \mathcal{L}_{\mathcal{V}} =  - \sum_{u=1}^{U} \textrm{var}_{k}\{\alpha_{u}^{k}\}
\end{equation}

We add these losses only on the $\mathcal{G}$ step of the \ac{bcd} training, to solely regularise the gating network. We ablate the proposed training schedule in Section \ref{sec:ablation}.

\subsection{Sign Language Output}

Generating a video from the produced skeleton pose sequence is a trivial task, connecting the relevant joints of each frame as seen in Figure \ref{fig:qual_skels}. 

\section{Experiments} \label{sec:experiments}

\paragraph{Implementation Details}

In our experiments, we build translation sub-networks with two layers, $2L$, two heads, $2H$, and embedding size of 256, $256Em$. \hmm{The architecture for motion primitives and gating network are $2L, 2H, 128Em$ and $2L, 4H, 64Em$, respectively.} We utilise a word embedding Our proposed architecture contains only 7.8M parameters, compared to 16.3M for the SOTA model \cite{saunders2021continuous}. We apply Gaussian noise with a noise rate of 5, as proposed by Saunders \etal \cite{saunders2020progressive}. All parts of our network are trained with Xavier initialisation \cite{glorot2010understanding}, Adam optimization \cite{kingma2014adam} with default parameters and a learning rate of $10^{-4}$ for the gating network and $10^{-3}$ for the rest. Our code is based on Kreutzer et al.'s NMT toolkit, JoeyNMT \cite{JoeyNMT}, and implemented using PyTorch \cite{paszke2017automatic}. 

\paragraph{Dataset} \label{sec:dataset}
We evaluate our approach on the publicly available \ac{ph14t} dataset introduced by Camgoz et al. \cite{camgoz2018neural}. The corpus provides parallel sequences of 8257 German sentences, sign gloss translations and sign pose videos. \hmm{This is a challenging dataset due to the low video quality. However, there are few other comprehensive sign language datasets available \cite{camgoz2021content4all}.} We train our \ac{momp} model to generate sign pose sequences of skeleton joint positions. Manual and non-manual features of each video are extracted in 2D using OpenPose \cite{cao2018openpose}, with the manuals lifted to 3D using the skeletal model estimation model proposed in \cite{zelinka2020neural}. We normalise the skeleton pose as in \cite{saunders2020progressive}.

\paragraph{Evaluation}

To compare against the state-of-the-art, we use the back translation evaluation metric \cite{saunders2020progressive}, which employs a pre-trained \ac{slt} model \cite{camgoz2020sign} to translate the produced sign pose sequences back to spoken language. BLEU and ROUGE scores are computed against the original input, with BLEU n-grams from 1 to 4 provided for completeness. The \ac{slp} evaluation protocols on the \ac{ph14t} dataset, set by \cite{saunders2020progressive}, are as follows: \textit{Gloss to Pose (G2P)} is the production of sign pose from gloss intermediary, evaluating the sign production capabilities; \textit{Text to Pose (T2P)} is the production of sign pose directly from spoken language, and is the more difficult end-to-end test of an \ac{slp} system.

\subsection{Quantitative Evaluation}

\paragraph{Number of Motion Primitives}

We start our experiments on the \textit{Gloss to Pose} task, and evaluate the production capabilities of the animation sub-network. We therefore set the translation loss, $\mathcal{L}_{\mathcal{T}}$, to zero. Our first experiment evaluates the performance when varying the number of motion primitive experts, $\mathcal{K}$. Although having a larger number of motion primitives allows each to be more specialised, it also makes the models harder to converge and prone to overfitting. To this end, we build \ac{momp} networks using 6 to 10 primitives and evaluated their \textit{Gloss to Pose} performance.

As shown in Table \ref{tab:motion_primitives}, we find that 8 motion primitives performs best, achieving a 13.32 BLEU-4 score on the development set. This gives a balance between specialisation of experts and training convergence difficulty, as we find that too many experts leads to an overfit. For the rest of our experiments, we constructed our \ac{momp} model with 8 motion primitives.

\begin{table}[t!]
\centering
\resizebox{0.9\linewidth}{!}{%
\begin{tabular}{@{}p{2cm}cc|cc@{}}
\toprule
\multicolumn{1}{c|}{\# of Motion}  & \multicolumn{2}{c}{DEV SET} & \multicolumn{2}{c}{TEST SET}  \\ 
\multicolumn{1}{c|}{Primitives:}  & BLEU-4 & ROUGE & BLEU-4 & ROUGE \\ \midrule
\multicolumn{1}{c|}{6} & 12.67 & 35.17 & 12.38 & 35.29 \\
\multicolumn{1}{c|}{7} & 12.57 & 35.90 & 12.15 & 35.37 \\
\multicolumn{1}{c|}{8} & \textbf{13.32} & \textbf{37.58} & \textbf{12.67} & \textbf{35.61} \\
\multicolumn{1}{c|}{9} & 12.55 & 36.14 & 12.31 & 34.93 \\
\multicolumn{1}{c|}{10} & 12.53 & 35.90 & 11.99 & 34.62 \\
\bottomrule
\end{tabular}%
}
\caption{Impact of different numbers of motion primitives on the performance of \ac{momp} for the \textit{Gloss to Pose} task.}
\label{tab:motion_primitives}
\end{table}

\paragraph{Ablation Study} \label{sec:ablation}

We next ablate our \ac{momp} network to highlight the importance of each proposed network attribute. Table \ref{tab:ablation} shows model performance on the \textit{Gloss to Pose} task. We first remove the randomness applied to the \ac{bcd} training, as described in section \ref{sec:bcd} (\hbox{\ac{momp} - $\mathrm{Rand}$}). Model performance is significantly degraded, resulting in 12.14 BLEU-4 on the development set. This is due to the removal of any ability for exploration in the expert update, $\mathcal{F}$, step of \ac{bcd}.

Removing \ac{bcd} training entirely (\hbox{\ac{momp} - $BCD$}) can be seen to negatively impact model performance further, to 10.85 BLEU-4. This is due to the combined update of both gating network and expert parameters leading to an unstable \ac{moe} model with non-specialised experts, as seen in previous works \cite{peng2020mixture,shen2019mixture}. We additionally conduct experiments with a simple $EM$ training (\hbox{\ac{momp} + $EM$}), alternating updates between the gating network, and a non-sampled combination of motion primitives. However, this still resulted in poor performance of 11.84 BLEU-4.

Removing the expert balance loss (\hbox{\ac{momp} - $\mathcal{L}_{\mathcal{B}}$}) results in an unbalanced gating network that activates only a single motion primitive. This means that the model does not take full advantage of the multiple experts available for specialisation, resulting in a poor performance of 12.20 BLEU-4. Removing the variance loss (\hbox{\ac{momp} - $\mathcal{L}_{\mathcal{V}}$}) causes each frame to have a combination of experts rather than a sparse representation. This results in a blended output which regresses to the mean, causing a non-expressive skeletal pose and poor performance of only 11.88 BLEU-4. 

% The impact of these losses can be further seen in the blending coefficient graphs in Figure \ref{fig:blending_coefficients}.

% This results in a poor performance of only 11.88 BLEU-4, due to the blending of expert outputs resulting in a mean, non-expressive output. The blending between experts was also not seen during the \ac{bcd} training schedule, causing a drastic difference at inference. 

% Figure \ref{fig:blending_coefficients} shows example of blending coefficients per frame when both $\mathcal{L}_{\mathcal{B}}$ and $\mathcal{L}_{\mathcal{V}}$ is applied.

% Overall, this ablation study highlights the importance of the proposed training schedules of \ac{bcd} training and expert balancing losses.

\begin{table}[t!]
\centering
\resizebox{0.9\linewidth}{!}{%
\begin{tabular}{@{}p{2cm}cc|cc@{}}
\toprule
  & \multicolumn{2}{c}{DEV SET} & \multicolumn{2}{c}{TEST SET}  \\ 
\multicolumn{1}{c|}{Approach:}  & BLEU-4 & ROUGE & BLEU-4 & ROUGE \\ \midrule
\multicolumn{1}{l|}{\textbf{\ac{momp}}} & \textbf{13.32} & \textbf{37.58} & \textbf{12.67} & \textbf{35.61}  \\
\multicolumn{1}{l|}{\ac{momp} - $\mathrm{Rand}$} & 12.14 & 35.67 & 11.93 & 35.45 \\
\multicolumn{1}{l|}{\ac{momp} - $BCD$} & 10.85 & 33.64 & 10.40 & 32.11 \\
\multicolumn{1}{l|}{\ac{momp} + $EM$} & 11.84 & 35.16 & 11.63 & 34.71 \\
\multicolumn{1}{l|}{\ac{momp} - $\mathcal{L}_{\mathcal{B}}$} & 12.20 & 35.43 & 11.72 & 34.60 \\
\multicolumn{1}{l|}{\ac{momp} - $\mathcal{L}_{\mathcal{V}}$} & 11.88 & 35.47 & 11.45 & 34.45 \\
\bottomrule
\end{tabular}%
}
\caption{Ablation study of \ac{momp} performance for the \textit{Gloss to Pose} task.}
\label{tab:ablation}
\end{table}

\begin{table}[b]
\centering
\resizebox{0.9\linewidth}{!}{%
\begin{tabular}{@{}p{1cm}p{1cm}cc|cc@{}}
\toprule
\multicolumn{2}{c|}{Loss Weights}     & \multicolumn{2}{c}{DEV SET} & \multicolumn{2}{c}{TEST SET} \\ 
\multicolumn{1}{c|}{$\lambda_{\mathcal{T}}$} & \multicolumn{1}{c|}{$\lambda_{\mathcal{A}}$}  & BLEU-4 & ROUGE & BLEU-4 & ROUGE \\ \midrule
\multicolumn{1}{c|}{0.0} & \multicolumn{1}{c|}{1.0} & 12.74 & 36.17 & 12.16 & 35.53 \\ \midrule
\multicolumn{1}{c|}{1.0} & \multicolumn{1}{c|}{1.0} & 13.72 & 37.63 & 13.18 & 36.84 \\
\multicolumn{1}{c|}{2.0} & \multicolumn{1}{c|}{1.0} & \textbf{14.03} & \textbf{37.76} & \textbf{13.30} & 36.77 \\
\multicolumn{1}{c|}{5.0} & \multicolumn{1}{c|}{1.0} & 13.69 & 37.67 & 13.12 & \textbf{37.10} \\
\multicolumn{1}{c|}{10.0} & \multicolumn{1}{c|}{1.0} & 13.51 & 36.99 & 12.83 & 36.53 \\
\bottomrule
\end{tabular}%
}
\caption{Impact of different translation and animation loss weightings on \ac{momp} \textit{Text to Pose} performance.}
\label{tab:translation_animation}
\end{table}

\begin{table*}[t!]
\centering
\resizebox{0.85\linewidth}{!}{%
\begin{tabular}{@{}p{2.8cm}ccccc|ccccc@{}}
\toprule
     & \multicolumn{5}{c}{DEV SET}  & \multicolumn{5}{c}{TEST SET} \\ 
\multicolumn{1}{c|}{Approach:}  & BLEU-4         & BLEU-3         & BLEU-2         & BLEU-1         & ROUGE          & BLEU-4         & BLEU-3         & BLEU-2         & BLEU-1         & ROUGE          \\ \midrule
\multicolumn{1}{r|}{Progressive Transformers \cite{saunders2020progressive}}    & 11.93 & 15.08 & 20.50 & 32.40 & 34.01 & 10.43  & 13.51 & 19.19 & 31.80 & 32.02 \\
\multicolumn{1}{r|}{Adversarial Training \cite{saunders2020adversarial}} & 13.16 & 16.52 & 22.42 & 34.09 & 36.75 & 12.16 & 15.31 & 20.95 & 32.41 & 34.19 \\
\multicolumn{1}{r|}{Mixture Density Networks \cite{saunders2021continuous}} & 13.14 & \textbf{16.77} & 22.59 & 33.84 & \textbf{39.06} & 11.94 & 15.22 & 21.19 & 33.66 & 35.19 \\
\multicolumn{1}{r|}{\textbf{\ac{momp}} \textbf{(Ours)}} & \textbf{13.32} & 16.71 & \textbf{22.67} & \textbf{34.21} & 37.58 & \textbf{12.67} & \textbf{16.03} & \textbf{22.02} & \textbf{33.95} & \textbf{35.61} \\
\bottomrule
\end{tabular}%
}
\caption{Back translation results on the \ac{ph14t} dataset for the \textit{Gloss to Pose} task.}
\label{tab:gloss_to_pose}
\end{table*}

\begin{table*}[t!]
\centering
\resizebox{0.85\linewidth}{!}{%
\begin{tabular}{@{}p{2.8cm}ccccc|ccccc@{}}
\toprule
     & \multicolumn{5}{c}{DEV SET}  & \multicolumn{5}{c}{TEST SET} \\ 
\multicolumn{1}{c|}{Approach:}  & BLEU-4         & BLEU-3         & BLEU-2         & BLEU-1         & ROUGE          & BLEU-4         & BLEU-3         & BLEU-2         & BLEU-1         & ROUGE          \\ \midrule
\multicolumn{1}{r|}{Progressive Transformers \cite{saunders2020progressive}}    & 11.82 & 14.80 & 19.97 & 31.41 & 33.18 & 10.51 & 13.54 & 19.04 & 31.36 & 32.46 \\ 
\multicolumn{1}{r|}{Adversarial Training \cite{saunders2020adversarial}} & 12.65 & 15.61 & 20.58 & 31.84 & 33.68 & 10.81 & 13.72 & 18.99 & 30.93 & 32.74 \\
\multicolumn{1}{r|}{Mixture Density Networks \cite{saunders2021continuous}} & 11.54 & 14.48 & 19.63 & 30.94 & 33.40 & 11.68 & 14.55 & 19.70 & 31.56 & 33.19 \\ 
\multicolumn{1}{r|}{\textbf{\ac{momp}} \textbf{(Ours)}} & \textbf{14.03} & \textbf{17.50} & \textbf{23.49} & \textbf{35.23} & \textbf{37.76} & \textbf{13.30} & \textbf{16.86} & \textbf{23.27} & \textbf{35.89} & \textbf{36.77} \\
\bottomrule
\end{tabular}%
}
\caption{Back translation results on the \ac{ph14t} dataset for the \textit{Text to Pose} task.}
\label{tab:text_to_pose}
\end{table*}

\paragraph{Translation and Animation} \label{sec:translation_animation_evaluation}

%% ************* READ THIS ************************ %%
% \todo{For the CRC, update the model performance based on the further parameter/architecture search done for BMVC}
%% ************* READ THIS ************************ %%

In our next set of experiments, we switch to the full \textit{Text to Pose} task. We examine the performance gain from adding the translation loss, $\mathcal{L}_{\mathcal{T}}$, alongside the animation loss, $\mathcal{L}_{\mathcal{A}}$. As a baseline, we trained a \ac{momp} model solely with the animation loss and with no gloss supervision, by setting the translation weight, $\lambda_{\mathcal{T}}$, to zero. We then jointly train for both translation and animation, with various weightings between the losses.

Table \ref{tab:translation_animation} shows experiments on the loss weightings, $\lambda_{T}$ and $\lambda_{A}$. As can be seen, jointly training \ac{momp} on both translation and animation with equal weightings (${\mathcal{L}_{\mathcal{T}}=\mathcal{L}_{\mathcal{A}}=1}$) significantly improves the back translation performance to 13.72 BLEU-4. This demonstrates the value of explicitly training on both the translation and animation sub-tasks. Increasing the translation loss weighting to $\mathcal{L}_{\mathcal{T}}=2$ further increased performance to 14.03 BLEU-4, with even larger translation losses degrading performance. We believe this is due to a required balance between gloss supervision and the ultimate \ac{slp} performance.

\begin{table}[b!]
\centering
\resizebox{0.8\linewidth}{!}{%
\begin{tabular}{@{}p{0.0cm}ccc@{}}
% \begin{tabular}{@{}p{0.0cm}p{1.25cm}p{1.25cm}p{1.25cm}p{1.25cm}@{}}
\toprule
%  & \multicolumn{3}{c}{Skeleton} & \multicolumn{3}{c}{Hand} \\ 
 & \multicolumn{1}{c}{Ours} & \multicolumn{1}{c}{Baseline \cite{saunders2021continuous}} & \multicolumn{1}{c}{No Pref.} \\ \midrule
\multicolumn{1}{r|}{Skeletons} & \textbf{49\%} & 33\% & 18\% \\
\multicolumn{1}{r|}{Hands} & \textbf{50\%} & 36\% & 14\%\\
\bottomrule
\end{tabular}%
}
\caption{Perceptual study results, showing the percentage of participants who preferred our outputs, the baseline output or had no preference, for both overall skeleton and hands.}
\label{tab:perceptual}
\end{table}

\paragraph{Baseline Comparison}

We compare the performance of \ac{momp} against 3 baseline \ac{slp} models: 1) Progressive transformers \cite{saunders2020progressive}, which applied the classical transformer architecture to sign language production. 2) Adversarial training \cite{saunders2020adversarial}, which utilised an adversarial discriminator to prompt more expressive productions and 3) \acp{mdn} \cite{saunders2021continuous}, which modelled the variation found in sign language using multiple distributions to parameterise the entire prediction subspace.

Table \ref{tab:gloss_to_pose} shows that \ac{momp} achieves state-of-the-art \textit{Gloss to Pose} results of 13.32/12.67 BLEU-4 for the development and test sets, respectively. This shows the expressive sign language sequences that can be produced from the animation sub-task, highlighting the effect of the learnt motion primitives within the proposed \ac{momp} network.

\textit{Text to Pose} results are shown in Table \ref{tab:text_to_pose}, with \ac{momp} achieving 14.03/13.30 BLEU-4 for the development and test sets respectively, an 11\% improvement over the state-of-the-art. These results highlight the significant success of detaching the translation and animation sub-tasks for the ultimate task of \ac{slp}. 

Furthermore, the performance in the \textit{Text to Pose} task is higher than the \textit{Gloss to Pose} task. This is surprising and significant for \ac{slp}, as \textit{Gloss to Pose} is often quoted as the simpler task. We believe this is due to the wider context available in spoken language compared to sign glosses. As we are not forcing translation to go via a gloss bottleneck, the model has access to more subtle grammatical cues for sign production. Instead, we are using the gloss information to supervise end-to-end training from spoken language. This is important for scaling \ac{slp} to domains that have limited gloss annotations, which can be expensive to obtain.

\paragraph{Perceptual Study}

We perform a perceptual study of our skeleton pose productions, showing participants pairs of videos produced by \ac{momp} and the state-of-the-art baseline \ac{mdn} model \cite{saunders2021continuous}. Participants were asked to select which video had the best life-like motion, for the overall skeleton and specifically the hands. In total, 24 participants completed the study, of which 13\% were signers. Table \ref{tab:perceptual} shows the percentage of participants who preferred our outputs, the baseline outputs or had no preference between them for both the overall skeleton and hands. 

It can be clearly seen that our outputs were preferred by participants compared to the baseline for both the overall skeleton (49\%) and specifically the hands (50\%), with only 33\% (skeleton) and 36\% (hands) preferring the baseline. This further suggests that the proposed \ac{momp} network produces expressive and life-like animations from the combinations of learnt motion primitives.

\subsection{Qualitative Evaluation}

\begin{figure*}[t!]
    \centering
    \includegraphics[width=0.86\linewidth]{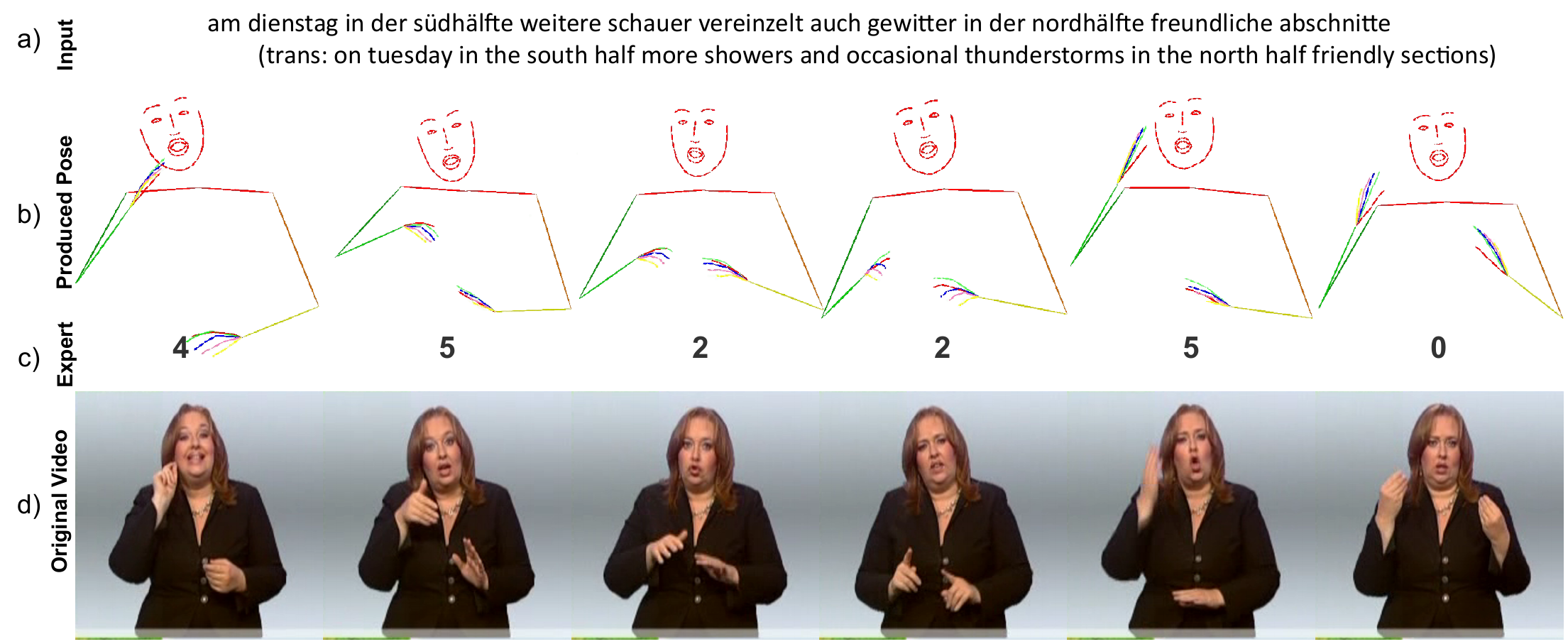}
    % \draftfig
    \caption{Qualitative results, showing a) Source spoken language, b) Produced sign pose sequence, c) Expert allocation per frame and d) Original video for comparison.}
    \label{fig:qual_skels}
\end{figure*}%

\begin{figure*}[b]
    \centering
    \begin{subfigure}[b]{0.3\textwidth}
     \centering
     \includegraphics[width=\textwidth]{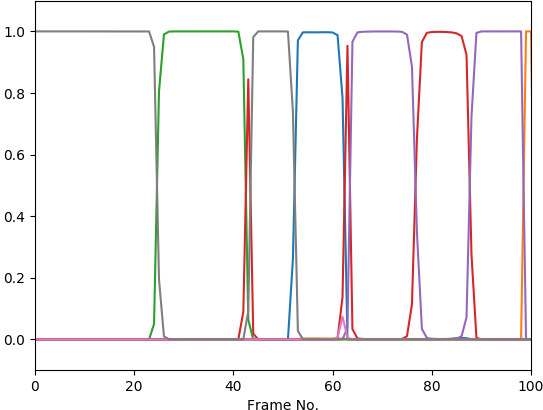}
    \end{subfigure}
    %  \hfill
    \begin{subfigure}[b]{0.3\textwidth}
     \centering
     \includegraphics[width=\textwidth]{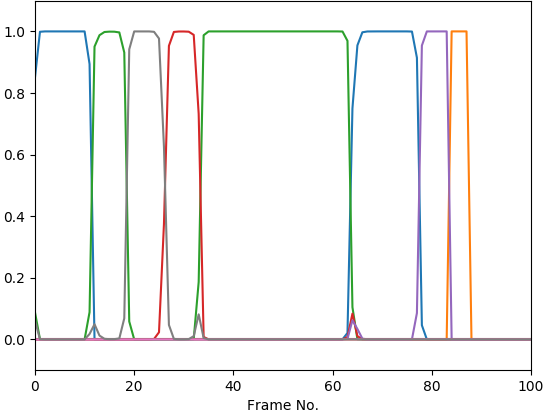}
    \end{subfigure}
    %  \hfill
    \begin{subfigure}[b]{0.3\textwidth}
     \centering
     \includegraphics[width=\textwidth]{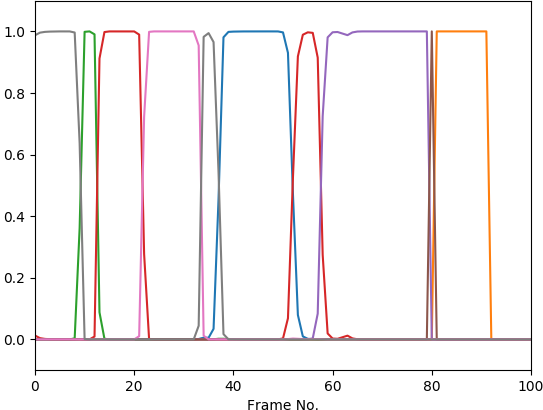}
    \end{subfigure}
    \caption{Example expert allocation graphs showing blending coefficient weight per expert against frame.}
    \label{fig:blending_coefficients}
\end{figure*}%

In this section we report qualitative results. Figure \ref{fig:qual_skels} shows signs automatically produced from a source spoken language sentence alongside the original video for comparison. The activated expert is shown for each produced skeleton, highlighting the usage of multiple motion primitives over a full sequence. Expert 5 can be seen to produce right handed motions, whereas expert 2 deals with downward motions of both hands.

\paragraph{Motion Primitives}

To evaluate the importance of each motion primitive at test time, we selectively deactivate a particular primitive and observe the visual results. Practically, we set the pre-softmax blending coefficient to minus infinity, to force $\alpha_{1:U}^{k} = 0$. This required the model to activate the other experts for these frames. 

We observe the output skeleton pose degrades to a mean pose when expert $k$ was intended to be activated, performing a non-expressive motion.  We notice a particular affect on the non-manual features that degrade to a significantly worse output. In addition, we find an average performance drop of 0.83 BLEU-4 when disabling a single motion primitive. We believe this phenomenon is due to expert $k$ becoming specialised for the desired motion, meaning all other experts were not trained to perform the motion. 

\paragraph{Blending Coefficient Graphs}

Figure \ref{fig:blending_coefficients} shows some example blending coefficients plotted per sequence. The graphs show the blending coefficients of each weight per frame of a sequence, with each expert plotted as a different colour. As seen, different motion primitives are activated for distinct sections of the sequences, combining unique motions to create a continuous sign language sequence.

The balanced nature of the experts can be seen, as each expert is represented over the sequences. This highlights the effect of the balancing loss, $\mathcal{L}_{\mathcal{B}}$, in ensuring the full repertoire of experts is exploited. In addition, each frame is represented by a single expert, showing the impact of the variance loss, $\mathcal{L}_{\mathcal{V}}$.

\section{Conclusion} \label{sec:conclusion}

Even though \ac{slp} requires both an accurate translation and expressive production, previous works have combined these tasks into a single end-to-end architecture with one unified loss function \cite{saunders2020progressive,stoll2020text2sign,zelinka2020neural}.

In this paper, we proposed separating the \ac{slp} task into two distinct jointly-trained sub-tasks. The first translation sub-task translates from spoken language to sign language representation, with explicit gloss supervision. Subsequently, an animation sub-task produces expressive sign language sequences that closely resemble the gloss representation. Motivated by phonetics, we proposed a \acf{momp} architecture, a novel \ac{moe} based network that learns to combine distinct motion primitives to produce a continuous sign language sequence.

We evaluated \ac{momp} on the \ac{ph14t} dataset, with perceptual studies showing that \ac{momp} achieves the best animation quality. We achieved state-of-the-art back translation performance, and reported better \ac{slp} performance for direct translation from text, i.e. \textit{Text to Pose}, compared to from gloss intermediaries, i.e. \textit{Gloss to Pose}.

%------------------------------------------------------------------------
\section{Acknowledgements}
This project was supported by the EPSRC project ExTOL (EP/R03298X/1), the SNSF project SMILE2 (CRSII5\_193686) and the EU project EASIER (ICT-57-2020-101016982). This work reflects only the authors view and the Commission is not responsible for any use that may be made of the information it contains. We would also like to thank NVIDIA Corporation for their GPU grant.

{\small
\bibliographystyle{ieee_fullname}
\bibliography{bibliography}
}

\end{document}

%% file: acronyms.tex
\begin{acronym}[PHOENIX14\textbf{T} ]

\acrodefplural{rnn}[RNNs]{Recurrent Neural Networks}
\acrodefplural{cnn}[CNNs]{Convolutional Neural Networks}
\acrodefplural{hmm}[HMMs]{Hidden Markov Models}
\acrodefplural{gru}[GRUs]{Gated Recurrent Units}
\acrodefplural{crf}[CRFs]{Conditional Random Fields}
\acrodefplural{gan}[GANs]{Generative Adversarial Networks}
\acrodefplural{gpu}[GPUs]{Graphic Processing Units}

\acrodefplural{mdn}[MDNs]{Mixture Density Networks}

\acro{bsl}[BSL]{British Sign Language}
\acro{bleu}[BLEU]{Bilingual Evaluation Understudy}
\acro{blstm}[BLSTM]{Bidirectional Long Short-Term Memory}
\acro{bcd}[BCD]{Block Coordinate Descent}
\acro{cnn}[CNN]{Convolutional Neural Network}
\acro{crf}[CRF]{Conditional Random Field}
\acro{cslr}[CSLR]{Continuous Sign Language Recognition}
\acro{ctc}[CTC]{Connectionist Temporal Classification}
\acro{dl}[DL]{Deep Learning}
\acro{dgs}[DGS]{German Sign Language - Deutsche Gebärdensprache}
\acro{dsgs}[DSGS]{Swiss German Sign Language - Deutschschweizer Geb\"ardensprache}
\acro{dtw}[DTW]{Dynamic Time Warping}
\acro{fc}[FC]{Fully Connected}
\acro{ff}[FF]{Feed Forward}
\acro{gan}[GAN]{Generative Adversarial Network}
\acro{gpu}[GPU]{Graphics Processing Unit}
\acro{gru}[GRU]{Gated Recurrent Unit}
\acro{gtpt}[G2PT]{Gloss to Pose Transformer}
\acro{hmm}[HMM]{Hidden Markov Model}
\acro{hpe}[HPE]{Hand Pose Enhancer}
\acro{isl}[ISL]{Irish Sign Language}
\acro{lstm}[LSTM]{Long Short-Term Memory}
\acro{mha}[MHA]{Multi-Headed Attention}
\acro{mtc}[MTC]{Monocular Total Capture}
\acro{mse}[MSE]{Mean Squared Error}
\acro{mdn}[MDN]{Mixture Density Network}
\acro{moe}[MoE]{Mixture-of-Expert}
\acro{momp}[\textsc{MoMP}]{Mixture of Motion Primitives}
\acro{nmt}[NMT]{Neural Machine Translation}
\acro{nlp}[NLP]{Natural Language Processing}
\acro{ph12}[PHOENIX12]{RWTH-PHOENIX-Weather-2012}
\acro{ph14}[PHOENIX14]{RWTH-PHOENIX-Weather-2014}
\acro{ph14t}[PHOENIX14\textbf{T}]{RWTH-PHOENIX-Weather-2014\textbf{T}}
\acro{pof}[POF]{Part Orientation Field}
\acro{paf}[PAF]{Part Affinity Field}
\acro{pttt}[P2TT]{Pose to Text Transformer}
\acro{relu}[RELU]{Rectified Linear Units}
\acro{rnn}[RNN]{Recurrent Neural Network}
\acro{rouge}[ROUGE]{Recall-Oriented Understudy for Gisting Evaluation}
\acro{sgd}[SGD]{Stochastic Gradient Descent}
\acro{sla}[SLA]{Sign Language Assessment}
\acro{slr}[SLR]{Sign Language Recognition}
\acro{slt}[SLT]{Sign Language Translation}
\acro{slp}[SLP]{Sign Language Production}
\acro{smt}[SMT]{Statistical Machine Translation}

\acro{ttgt}[T2GT]{Text to Gloss Transformer}
\acro{ttpt}[T2PT]{Text to Pose Transformer}
\acro{ttp}[T2P]{Text to Pose}
\acro{ttgtp}[T2G2P]{Text to Gloss to Pose}
\acro{wer}[WER]{Word Error Rate}

\end{acronym}